\documentclass[review]{elsarticle}
\usepackage[utf8]{inputenc}
\usepackage{gensymb}
\usepackage{graphicx}
\usepackage{xcolor}
\usepackage{subfig}
\usepackage{flafter} 
\usepackage{float}
\usepackage{siunitx}
\usepackage{multirow}
\usepackage{booktabs}
\usepackage{amsmath,bm}
\usepackage[ruled,vlined]{algorithm2e}
\usepackage{multirow}
\usepackage{comment}
\journal{Information Fusion}

\renewcommand{\vec}[1]{\bm{#1}}
\title{Towards a More Reliable Interpretation of Machine Learning Outputs for Safety-Critical Systems using Feature Importance Fusion}

\author[mymainaddress]{Divish Rengasamy\corref{mycorrespondingauthor}}
\cortext[mycorrespondingauthor]{Corresponding author}
\ead{divish.rengasamy@nottingham.ac.uk}

\author[mymainaddress]{Benjamin Rothwell}
\author[mysecondaryaddress,mythirdaddress]{Grazziela Figueredo}

\address[mymainaddress]{Gas Turbine and Transmissions Research Centre, The University of Nottingham, UK}
\address[mysecondaryaddress]{Advanced Data Analysis Centre, The University of Nottingham, UK}
\address[mythirdaddress]{School of Computer Science, The University of Nottingham, UK}

\begin{document}

\begin{frontmatter}
%\maketitle

\begin{abstract}

When machine learning supports decision-making in safety-critical systems, it is important to verify and understand the reasons why a particular output is produced. Although feature importance calculation approaches assist in interpretation, there is a lack of consensus regarding how features' importance is quantified, which makes the explanations offered for the outcomes mostly unreliable. A possible solution to address the lack of agreement is to combine the results from multiple feature importance quantifiers to reduce the variance of estimates. Our hypothesis is that this will lead to more robust and trustworthy interpretations of the contribution of each feature to machine learning predictions. To assist test this hypothesis, we propose an extensible Framework divided in four main parts: (i) traditional data pre-processing and preparation for predictive machine learning models; (ii) predictive machine learning; (iii) feature importance quantification and (iv) feature importance decision fusion using an ensemble strategy. We also introduce a novel fusion metric and compare it to the state-of-the-art. Our approach is tested on synthetic data, where the ground truth is known. We compare different fusion approaches and their results for both training and test sets.  We also investigate how different characteristics within the datasets affect the feature importance ensembles studied. Results show that our feature importance ensemble Framework overall produces 15\% less feature importance error compared to existing methods. Additionally, results reveal that different levels of noise in the datasets do not affect the feature importance ensembles\ '  ability to accurately quantify feature importance, whereas the feature importance quantification error increases with the number of features and number of orthogonal informative features. 
\end{abstract}

\end{frontmatter}

\clearpage

\section{Introduction}
\label{sec:introduction}

Extensive advances in Machine Learning (ML) have demonstrated its potential to successfully address complex problems in safety-critical areas, such as in healthcare~\cite{uddin2020body, miotto2018deep, cruz2017accurate}, aerospace~\cite{Rengasamy2018review,Rengasamy2020dynamic, yang2013data}, driver distraction~\cite{Mafeni2020distract, Eraqi2019drive, Mafeni2020cap}, civil engineering~\cite{farrar2012structural,catbas2016machine}, and manufacturing~\cite{zhang2019process, wang2018deep}. Historically, however, many ML models, specially those involving neural networks are viewed as `black boxes', where little is known about how the decision-making process takes place. The lack of adequate interpretability and verification of ML models~\cite{seshia2016towards, brundage2020toward} has therefore prevented an even wider adoption and integration of those approaches in high-integrity systems. For those domains, where mistakes are unacceptable due to safety, security and economical issues they may cause, the need to accurately interpret the predictions and inference of ML models becomes imperative. 

The recent rise in complexity of ML architectures has made it more difficult to explain their outputs. Although there is an overall agreement about the safety and ethical needs for interpreting ML outputs~\cite{brundage2020toward, pham2018asilomar}, there is however no consensus on how this challenge can be addressed. On one hand, there are advocates for the development of models that are themselves interpretable, rather than putting the effort later in making black-box models explainable~\cite{rudin2019stop}. The argument is that for critical decision making, explanations of the black-box models are often unreliable, can be misleading and therefore unsafe. Conversely, other researchers have focused their efforts into explaining complex ML models; and significant advances have been achieved~\cite{adadi2018peeking, gunning2017explainable,arrieta2020explainable}. 

An important approach to ML output elucidation adopting post-training explanation is given by feature importance estimates~\cite{chakraborty2017interpretability, altmann2010permutation}. There are multiple methods for calculating feature importance and they do not necessarily agree on how a feature relevance is quantified. It is not easy therefore to validate estimated feature importance, unless the ground truth is known. Furthermore, there is no consensus regarding  which is the best metric for feature importance calculation. 

The lack of consensus of current approaches in determining the importance of data attributes for ML decision making is a problem for safety-critical systems, as the explanation offered for the outcomes obtained is likely to be unreliable. There is therefore the need for more reliable and accurate ways of establishing feature importance. One possible strategy is to combine the results of multiple feature importance quantifiers, as a way to reduce the variance of estimates, leading to more a more robust and trustworthy interpretation of the contribution of each feature to the final ML model prediction. In this paper, we propose a general, adaptable and extensible Framework, which we have named as Multi-Method Ensemble (MME) for feature importance fusion with the objective of reducing variance in current feature importance estimates. The MME Framework is divided in four main parts: 

\begin{enumerate}
\item  The application of traditional data pre-processing and preparation approaches for computational modelling; 
\item Predictive modelling using ML approaches, such as random forest (RF), gradient boosted trees (GBT), support vector machines (SVM) and deep neural networks (DNN); 
\item Feature importance calculation for the ML models adopted, including values obtained from Permutation Importance (PI) \cite{strobl2007bias}, SHapley Additive exPlanations (SHAP)~\cite{lundberg2017unified} and Integrated Gradient (IG) \cite{sundararajan2017axiomatic}; 
\item Feature importance fusion strategy using ensembles. 
\end{enumerate}

For part 4, we also introduce a fusion metric, namely {\bf R}ank correlation with m{\bf A}jority vo{\bf TE} (RATE) and compare its performance to existing ensemble methods from the literature. The Framework as well as the ensemble strategies are tested on datasets considering different noise levels, number of important features and number of total features. In order to make sure our results are reliable, we conduct our experiments on synthetic data, where we determine beforehand the features and their importance. We compare the performance of our Framework to the more common method that uses a single feature importance method with multiple ML models to ensemble feature importance, which we named as Single Method Ensemble (SME). Additionally, we contrast and evaluate feature importance obtained from both training and test datasets to assess how sensitive ensemble feature importance determination methods are to the sets investigated and how much agreement is achieved. We also explore how different characteristics within the data affect feature importance ensembles. 

The remainder of the paper is structured as follows: Section 2 provides a background on different feature importance technique and literature review on the current state of ensemble feature importance techniques. Section 3 then introduces the methodology of how the dataset is generated and the proposed ensemble technique. Section 4 presents the results along with discussions on how different dataset affect the feature importance techniques and the performance of proposed ensemble feature importance. Finally, in Section 5, the conclusions and potential future work are suggested.

\section{Background}
\label{sec:background}

Early approaches to feature importance quantification utilise interpretable models, such as linear regression~\cite{nguyen2011linear} and logistic regression~\cite{shevade2003simple}; or ensembles, such as generalised linear models (GLM)~\cite{song2013random}, and decision trees (DT)~\cite{breiman2001random} to determine how each feature contributes to the model's output~\cite{kira1992practical}. As data problems become more  challenging and convoluted, simpler and interpretable models need to be replaced by complex ML solutions. For those, the ability to interpret predictions without the use of additional tools becomes far more difficult. Model-agnostic interpretation methods are commonly used strategies to help determining the feature importance from complex ML models. They are a class of techniques that determine feature importance, while treating models as black-box functions.  Our objective in this paper is to propose a Framework for the ensemble using these tools. In this section we review the current literature on ensemble feature importance, including the the basic concepts and rationale for choosing model-agnostic approaches in our Framework experiments.

\subsection{Related Work}
\label{sec:lit-review}

One of the earliest ensemble techniques to calculate feature importance, Random Forest (RF), propose by Breiman~\cite{breiman2001random}. RF is a ML model that forms from an ensemble of decision trees via random subspace methods~\cite{ho1998random}. Besides prediction, RF computes the overall feature importance by averaging those determined by each decision tree in the ensemble. RF feature importance is quantified depending on how many times a feature branches out in the decision tree, based on the Gini impurity metric. Alternatively, decision trees also calculate feature importance as Mean Decrease Accuracy (MDA) or more commonly known as PI by permutating the subset of features in each decision tree and calculating how much accuracy decreases as a consequence. Using the knowledge of ensembling feature importance from weak learners, De Bock {\textit et al. }~\cite{ debock2012gamens} proposes an ensemble learning based on generalised additive models (GAM) to estimate feature importance and confidence interval of prediction output. 
Similarly to Bagging, the average of each weaker additive model generates the ensemble predictions. The feature importance scores are generated using the following steps: (i) Generate output and calculate performance for individual predictions based on a specific performance criterion; (ii) permutate each feature and recalculate error for OOB predictions; (iii) calculate the partial importance score based on OOB predictions; (iv) Repeat step (i) to (iii) for each additive model and different forms of evaluation. The authors argue that the importance of each feature should be optimised according to the performance criteria most relevant to feature to obtain the most accurate feature importance score. The GAM ensemble-based feature importance is subsequently applied to identify essential features in churn prediction to determine customers likely to stop paying for particular goods and services. To determine the ten most relevant features, the authors use Receiver Operating Characteristic and top-decile lift. The authors observed that the sets of important features overlapped, but their rank order was different when using ROC and lift. The different rank orders show that feature importance is affected by the evaluation criteria. Both Breiman and De Bock {\textit et al. } uses only a single ML model with one type of feature importance method to calculate the ensemble feature importance, which restricts the potential to improve accuracy and to reduce variance. To overcome this limitation, Zhai and Chen~\cite{Zhai2018stacked} employ a stacked ensemble model to forecast the daily average of air particle concentrations in China. The stacked ensemble consists of four different ML models, namely, Least Absolute Shrinkage and Selection Operator (LASSO), adaptive boosting (AdaBoost), XGBoost, and Multi-layer Perceptron with Support Vector Regression (SVR) as the meta-regressor. The authors use a combination of feature selection and model generated method to determine feature importance, which is determined from Stability Feature Selections, XGBoost model and AdaBoost model. Their outputs are subsequently averaged for the final ranking of features. AdaBoost and GBT use the Mean Decrease Impurity (MDI) based on the Gini importance; SFS is based on maximum feature scores using Bayesian Information Criterion~\cite{schwarz1978estimating}. The top ten features are selected for evaluation. 

While Zhai and Chen used multiple ML models and one feature importance approach, to the best of our knowledge, there has not been further investigations to improve feature importance quantification using multiple models and multiple feature importance methods. Finally, as we can see in this section, there are minimal in-depth systematic investigation of how ensemble feature importance fusion works. Therefore, it is imperative that we investigate interpretability methods and  ensemble feature importance fusion under different data conditions.

\subsection{Permutation Importance}

PI measures feature importance by calculating the changes in model's error when a feature is replaced by a shuffled version of itself. The algorithm of how PI quantifies feature importance is as follows:

\begin{algorithm}[H]
\SetAlgoLined
\KwResult{Permutation feature importance}
\SetKwInOut{Input}{Input}
\Input{features\_array, labels, Trained\_Model}

predicted\_output = Trained\_Model(features))\;
baseline\_performance = Loss(Predicted\_output, labels)\;

\For{i = 0; i $<$ length(features); i++}{
    original\_feature = features[i]\;
    shuffled\_feature = shuffle(features[i])\;
    features[i] = shuffled\_feature\;
    predicted\_output = Trained\_Model(features))\;
    error = Loss(Predicted\_output, labels)\;
    feature\_importance[i] = error - baseline\_performance\;
    features[i] = original\_feature\;
 }
return feature\_importance\;
\caption{Algorithms of permutation importance}
\label{algo:pi}
\end{algorithm}
\bigskip
The magnitude of difference between $baseline\_performance$ and $error$ in Algorithm~\ref{algo:pi} signifies the importance of a feature. A feature has high importance if the performance of ML deviate significantly from the baseline after a shuffling; it therefore has low importance if the performance does not change significantly. PI can be run on train or test data but test data is usually chosen to avoid retraining of ML models to save computational overhead. If the computational cost is not an important factor to be considered, a drop-feature importance approach can be adopted to achieve greater accuracy. This is because in PI there is a possibility where a shuffled feature does not differ much from unshuffled feature for some instances. In contrast drop-feature importance excludes a feature, as oppose to performing its permutation. However, as it requires the ML model to be retrained every time a different features is dropped, it is computationally expensive for high dimensional data. In this paper we use PI over drop-feature importance to reduce the computational cost. 

Additionally, we choose to use PI over MDI using Gini Importance. MDI tends to disproportionately increase the importance of continuous or high-cardinality categorical variables~\cite{strobl2007bias}, leading to bias and unreliable feature importance measurements. PI is a model-agnostic approach to feature importance, and it can be used on GBT, RF, and DNN.

\subsection{Shapley Additive Explanations}

\begin{figure}[H]
    \centering
    \includegraphics[width=0.5\textwidth]{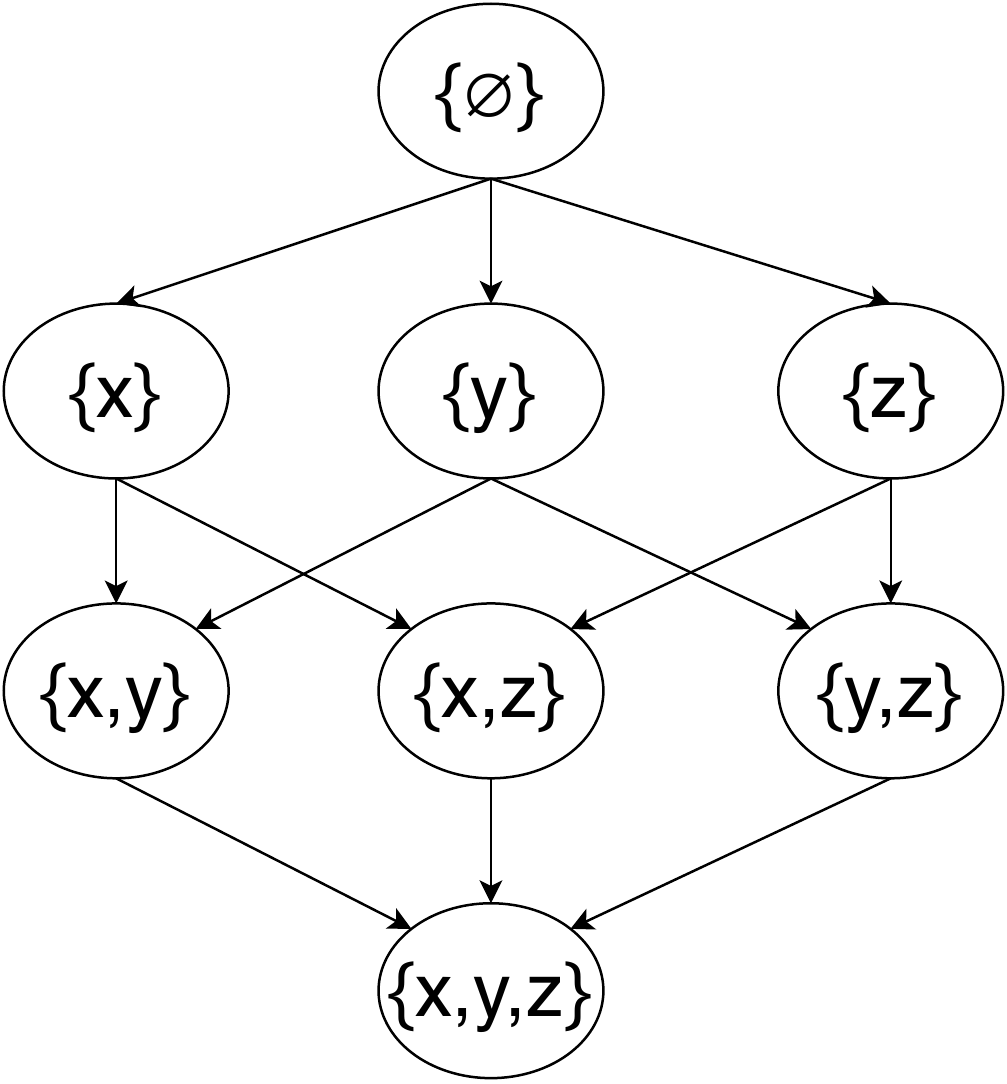}
    \caption{A graph representation of power set for features $\{x,y,z\}$. The $\o$ symbol represents the null set which is the average of all output. Each vertex represent a possible combination of features, and the edge shows the addition of new features previously not included in previous group of features.}
    \label{fig:power-set}
\end{figure}

Other widespread feature importance model-agnostic approaches in addition to PI and MDA is SHAP. SHAP is a ML interpretability method that uses Shapley values, a concept originally introduced by Lloyd Shapley~\cite{shapley1953value} in game theory to solve the problem of establishing each player contribution in cooperative games. Essentially, given a certain game scenario, the Shapley value is the average expected marginal contribution (MC) of a player after all possible combinations have been considered. For ML, SHAP determines the contribution of the available features of the model by assessing their every possible combination and quantifying their importance. 
The total possible combinations can be represented through a power set. For example, in the case of three features, $PowerSet\{x, y, z\}$ the power set is $\{\{\O\}, \{x\}, \{y\}, \{z\}, \{x,y\}, \{x,z\}, \{y,z\}, \{x,y,z\}\}$. Furthermore, Figure~\ref{fig:power-set} illustrates the relationships between the elements in the power set.

SHAP trains a ML model for each of the vertices shown in Figure~\ref{fig:power-set}. Therefore, there are $2^{\text{number of features}} = 2^{3} = 8$ models trained to estimate the contribution of each feature. The number of models needed to estimate feature importance using SHAP increases exponentially with the number of features. However, there are tools, such as the \textit{Python} library \textit{SHAP}~\cite{lundberg2020local2global} to accelerate the process through approximations and sampling. The MC of a feature can be calculated by traversing the graph in Figure~\ref{fig:power-set} and summing up the changes in output where the feature is previously absent from the combinations. For example, to calculate the contribution of feature $\{x\}$, we can calculate the weighted average of the change in the output from $\{\O\}$ to $\{x\}$, $\{y\}$ to $\{x,y\}$, $\{z\}$ to $\{x,z\}$, and $\{y,z\}$ to $\{x,y,z\}$. The MC of a feature $x$ going from $\{\O\}$ to $\{x\}$ is as follows:
\begin{equation}
    MC_{x,(\O,x)}(i_{0}) = Output_{x}(i_{0}) - Output_{\o}(i_{0})
\label{eq:mc}
\end{equation}

and following Equation~\ref{eq:mc} we calculate the SHAP value of feature $x$ of an instance as follows:

\begin{equation}
\begin{split} 
     SHAP_{x}(i_{0}) = w_{1}&*MC_{x,(\O,x)}(i_{0}) + \\
       w_{2}&*MC_{x,(x,y)}(i_{0}) + \\
       w_{3}&*MC_{x,(x,z)}(i_{0}) + \\
       w_{4}&*MC_{x,(x,y,z)}(i_{0}) + \\   
\label{eq:mc-instance}
\end{split}  
\end{equation}

The process is repeated for each feature to obtain the feature importance. The weights ($w_{1}, w_{2}, w_{3}, w_{4}$) in Equation~\ref{eq:mc-instance} sum to 1. The weights are calculated by taking the reciprocals of the number of possible combinations of MC for each row in Figure~\ref{fig:power-set}. For example, the weight of $w_{1}$ is ${3 \choose 1}^{-1}$. To calculate the global importance of a feature, the absolute SHAP values are averaged across all instances. 

Another approach is model-agnostic feature importance method is Local interpretable model-agnostic explanations (LIME)~\cite{ribeiro2016should}. We did not choose LIME as a model interpretation method because it does not have a good global approximation of feature importance and it is only able to provide feature importance for individual instances. Furthermore, LIME is sensitive to small perturbations in the input leading to different feature importance for similar input data~\cite{alvarez2018robustness}. 

\subsection{Integrated Gradients}
\label{sec:ig}

%There are several feature importance methods specifically designed for Deep Learning models, such as GradCAM, Guided GradCAM, Guided Backpropagation, DeepLift, and IG. However, not all of them are suitable for our investigations. GradCAM and Guided GradCAM, for instance, are only applicable to Convolutional Neural Networks (CNNs). Guided Backpropagation fails to deliver reliable feature importance calculation, as their  outputs remain invariant when the network is reparamaterised or when the test labels are randomly permutated~\cite{adebayo2018sanity}. Furthermore, a desirable property of feature importance is \textit{Completeness} which DeepLIFT and IG are designed to satisfy. \textit{Completeness} property states that the sum of feature importance should be the difference between the model's output at a particular instance and the baseline. While DeepLIFT is originally designed to satisfy the \textit{Completeness} property, it has since been shown that it fails to achieve that~\cite{ancona2017towards}.

IG is a gradient-based method for feature importance. It determines the feature importance $A$ in deep learning models by calculating the change in output, $f(x)$ relative to the change in input $x$. Additionally, the change in input features is approximated using an information-less baseline, $b$. The baseline input is a vector of zero in the case of regression to ensure that the baseline prediction is neutral and it functions as a counterfactual. The features importance are denoted by the difference between the characteristics of the deep learning model's output when features and baseline are used. The formula for feature importance using a baseline is as follows:

\begin{equation}
A_{i}^{f}(x,b) = f(x_{i}) - f(x[x_{i} = b_{i}])
\label{eq:att_base}
\end{equation}

\noindent
The individual feature is denoted by the subscript $i$. Equation~\ref{eq:att_base} can be also written in the form of gradient-based importance as:

\begin{equation}
G_{i}^{f}(x,b) = (x_{i} - b_{i})\frac{\partial f(x)}{\partial x_{i}}
\end{equation}

\noindent
IG obtains feature importance values by accumulating gradients of the features interpolated between the baseline value and the input. To interpolate between the baseline and the input, a constant, $\alpha$, with the value ranging from zero to one is used as follows:

\begin{equation}
IG_{i}^{f}(x,b) = (x_{i} - b_{i})\int_{\alpha=0}^{1} \frac{\partial f(b + \alpha(x-b))}{\partial x_{i}}d\alpha
\label{eq:ig}
\end{equation}

\noindent
Equation~\ref{eq:ig} is the final form of IG used to calculate feature importance in a deep learning model.

\section{The Proposed Feature Importance Fusion Framework}

This section introduces our proposed MME Framework where we employ multiple ML models and feature importance methods. We also introduce the different ensemble strategies investigated, including median, mode, box-whiskers, Modified Thompson Tau test, and majority vote. We also present the rank correlation with majority vote (RATE) fusion as a new ensembling strategy.

\subsection{Ensemble Feature Importance}
\begin{figure}[H]
    \centering
    \includegraphics[width=1.0\textwidth]{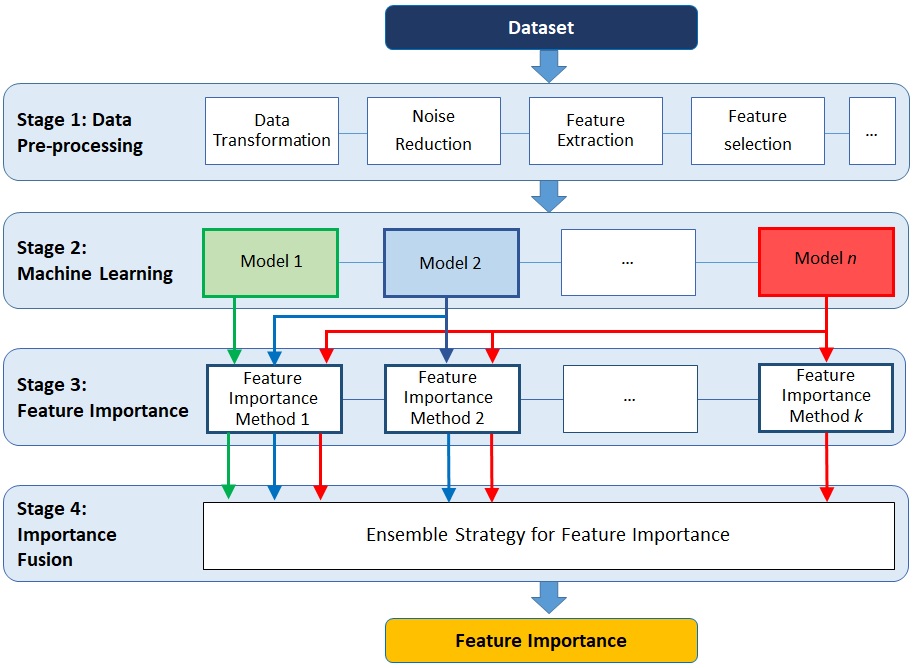}
    \caption{The four stages of the proposed feature importance fusion MME Framework. The first stage pre-processes the data and in the second step trains the data on multiple ML models. The third step calculate feature importance from the each trained ML models using multiple feature importance methods. Finally, the fourth step fuses all feature importance generated from the third step using an ensemble strategy to generate the final feature importance values.}
    \label{fig:ensemble-flow}
\end{figure}

 Figure~\ref{fig:ensemble-flow} shows our MME Framework. On {\bf Stage 1}, data undergoes pre-processing, such as transformation, noise reduction, feature extraction and feature selection. This stage is required for our Framework, as we need to ensure that the data has no inconsistencies and that the features used to train the machine learning models are orthogonal. The preference for features with low correlation guarantees that the feature importance calculation does not attribute random values of importance because a set of features contains similar information. On {\bf Stage 2} ML models are applied to the pre-processed data; subsequently, model-agnostics feature importance methods are applied to the ML models ({\bf Stage 3}). The feature importance results are fused into a final feature importance in {\bf Stage 4}. 
 
In {\bf Stage 3} and {\bf Stage 4} when multiple methods and ML models are employed in ensemble feature importance, we obtain several vectors, $V$ of feature importance values, $A$ as follows: $V = [A_{1}, A_{2}, ..., A{i}]$ where $i$ is the number of features. We need to establish a final decision from those vectors. So the first thing is to reduce the noise and anomalies in the decision. The feature importance vectors can be denoted as follows: $\vec{V} =[V_{1}, V_{2}, ..., V_{N}]$ where $N$ is the sum of each model multiplied by the number of feature importance methods used. To reduce the noise and variance, we can take the average of all feature importance vectors, $\overline{\vec{V}}$ which produces the variance, $\sigma^{2}/N$. As $N$ increases the variance decreases. The variance and correctness of final feature importance vector can be further improved if the anomalies are removed prior to taking the average.

\subsection{Ensemble Strategies}

Within our MME Framework, the importance calculated is stored in a matrix, $\overline{\Vec{V}}$, and the ensemble strategy, which can be obtained in several ways, is used to determine the final feature importance values from $\overline{\Vec{V}}$. The most common ensemble strategy, as discussed in Section~\ref{sec:lit-review} is to use the average values. However, this is not the most suitable ensemble strategy in cases where one or more of the feature importance approaches produce outliers compared to the majority of responses. So in addition to the mean, we also investigate data fusion using median, mode, box-whiskers, majority vote, Modified Thompson Tau test, and our novel fusion approach, RATE. For majority vote, each vector in the feature importance matrix have their features ranked based on their importance. Subsequently, the final feature importance is the average of the most common rank order for each feature. For example, feature $X_{i}$ has a final rank vector of $[1,1,1,2]$, where each rank $r_k$ is established by a different feature importance method $k$. The final feature importance value for feature $X_{i}$ is the average value from the three feature importance methods that ranked it as one. Modified Thompson Tau test is a statistical anomaly detection method using t-test to eliminate values that are above two standard deviations. 

\begin{figure}[H]
    \centering
    \includegraphics[width=1.0\textwidth]{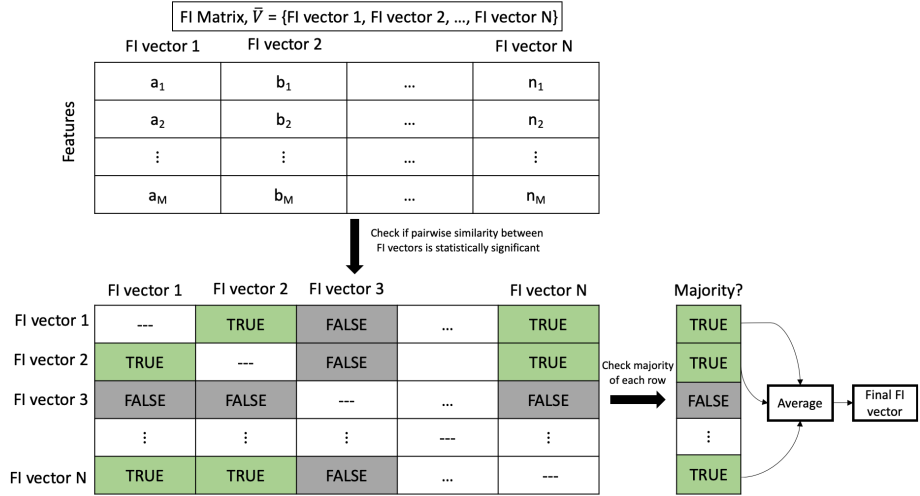}
    \caption{The working of RATE feature importance ensemble strategy. The feature importance (FI) vectors undergoes a rank correlation pairwise comparison to determine if the similarity between FI vectors is statistically significant (p-value$<$0.05). A value of `TRUE' is assigned if the two vectors are similar, otherwise, a `FALSE' value is assigned in a truth table. Each row of the truth table will go through a majority vote to determine if the FI vector is accounted when calculating the final FI.}
    \label{fig:rate}
\end{figure}

RATE is our novel fusion approach that combines the statistical test, feature rank and majority vote. RATE combines the advantage of using a statistical approach to rank feature importance and anomaly removal with majority vote. The steps used in RATE are illustrated in Figure~\ref{fig:rate}. The input to RATE is the feature importance matrix, $\overline{\vec{V}}$. $\overline{\vec{V}}$ is the matrix that has the individual feature importance from different models and it has the shape of $N*M$ where $N$ are the feature importance vectors from different importance calculation methods and $M$ is the number of features. We calculate the pairwise rank correlation between each feature importance vectors in the matrix $\overline{\vec{V}}$ to obtain the rank and the general correlation coefficient values~\cite{Kendall1948aa}. Using the rank and correlation value we determine whether the correlation between the vectors is statistically significant (p-value less than 0.05). The p-values are stored in a separate matrix that is converted to a truth table. If pairwise p-value is less than 0.05 it is given a value of `TRUE', otherwise it is `FALSE'. Using the truth table along with majority vote we determine which feature importance vector overall does not correlate with the majority of vectors and should be discarded. The remaining feature importance vectors are averaged as the final feature importance.

% \begin{algorithm}[H]
% \SetAlgoLined
% \SetKwInOut{Output}{Output}
% \Output{Feature importance}
% \SetKwInOut{Input}{Input}
% \Input{$\overline{\vec{V}}$, PairwiseRankPValue}
% \tcp{$\overline{\vec{V}}$ is the matrix of different feature importance vectors}
% \textcolor{red}{do the same and explain what PairwiseRankPValue is. As you added it as an input, it doesnt make sense to calculate here in the pseudo-code. So I removed it}
% \For{i,j $\in$ length(p\_values)}{
%     \tcp{p\_values $<$ 0.05 means that vectors are correlated}
%     \eIf{p\_values[i][j] $<$ 0.05}{ 
%         truth\_table[i][j] = TRUE\;
%         }{
%         truth\_table[i][j] = FALSE\;
%         }
% }
%   \For{i $\in$ row\_length(truth\_table)}{
%   \tcp{if feature importance method agrees with the majority of methods add it for final feature importance calculation}
%     \eIf{majority\_vote\_counter(truth\_table[i]) is TRUE}{
%         counter += 1\;
%         correlated\_method.append($\overline{\vec{V}}$[i])\;
%     }
%   }

% final\_feature\_importance = mean(correlated\_method)\;
% return final\_feature\_importance

%  \caption{RATE}
%  \label{algo:rank}
% \end{algorithm}

\section{Experimental Design}
\label{sec:methodology}

This section introduces our experimental design, including the benchmark datasets and how their generation is modulated. Subsequently, we discuss the data pre-processing, ML models employed, the evaluation metrics used and the experiments conducted.

\subsection{Data Generation}
The data investigated are generated using \textit{Python}'s\textit{ Scikit-learn} library~\cite{scikitlearn2011} with different characteristics to mimic a variety of real-world scenarios. The parameters used to modulate the creation of the datasets are the standard deviation of the Gaussian distributed noise applied to the data, the number of features included and the percentage of informative features. Their values are shown in Table~\ref{tab:data-params}.  We add Gaussian distributed noise with different standard deviation to the output as it has a more significant effect on prediction accuracy than that in the features~\cite{twala2013impact}. Although noise increases ML models' estimated error~\cite{twala2013impact, kalapanidas2003machine} studies investigating the relationship between data noise and feature importance error are scarce. Similarly, the impact of the number of features and how many features within the set are relevant to importance error is unknown.
A combination of values from each parameter in the table forms a dataset, and permutations of those parameters form a total of 45 datasets. For each dataset, we conduct ten experimental runs to ensure that the results are stable and reliable.

\begin{table}[h]
    \centering
    \begin{tabular}{lp{5cm}p{2cm}}
        \toprule
        Parameters & Description & Parameters' value \\
        \midrule
        Noise & Standard deviation of Gaussian noise applied to the output. & 0, 2, 4 \\
        Informative level (\%) & Percentage of informative features. Non-informative features do not contribute to the output. & 20, 40, 60, 80, 100 \\
        Number of features & Total number of features used to generate output values. & 20, 60, 100\\
        \bottomrule
    \end{tabular}
    \caption{Parameters to generate the datasets used to test our MME Framework.}
    \label{tab:data-params}
\end{table}

% Figure~\ref{fig:data-dist} illustrates the dataset features combination and the overall parameter space \textcolor{blue}{what is this figure adding to your paper? what do the numbers mean? you need a simpler diagram that adds to your story. this one is just showing a colourful diagram of permutation.}.

% \begin{figure}[H]
%     \centering
%     \includegraphics[width=1.0\textwidth]{Images/data_dist.png}
%     \caption{A slankey diagram to illustrate how the factors: noise level, informative level, and number of features combined to 45 distinct datasets.}
%     \label{fig:data-dist}
% \end{figure}

The data preprocessing in our case consists of normalising the input data to the range between zero and one. Normalisation accelerates the learning process and reduces model error for neural networks~\cite{sola1997importance}. Additionally, it allows for equal weighting of all features and therefore reduces bias during learning. 

\subsection{Machine Learning Models}
\label{sec:ml-models}
The ML approaches employed in our experiments are RF, GBT, and DNN; the hyperparameters adopted are shown in Table~\ref{table:hyperparams}.  We optimise the models using random hyperparameters search~\cite{bergstra2012random}, using conditions at the upper and lower limits of the parameters used for the datasets generation. The models are not optimised for individual datasets. We keep the model's hyperparameters constant as it might be a factor that affects feature importance accuracy. We therefore limit our objectives to investigating how data characteristics affect interpretability methods and how the appropriate fusion of different feature importance methods produces less biased estimates. Furthermore, we try to minimise overfitting in the deep learning model by adopting dropout~\cite{srivastava2014dropout} and L2 kernel regularisation~\cite{kukavcka2017regularization}.

\begin{table}[htpb]
    \centering
    \begin{tabular}{lp{6cm}p{4cm}}
        \toprule
        Models & Hyperparameters & Values \\
        \midrule
        Random Forest & Number of trees & 700\\
        & Maximum depth of trees & 7 levels\\
        & Minimum samples before split & 2\\
        & Maximum features & $\sqrt{p}$\\
        & Bootstrap & True\\
        \hline
        Gradient Boosted Trees & Number of trees & 700\\
        & Learning rate & 0.1\\
        & Maximum depth of trees & 7 levels\\
        & Loss function & Least square\\
        & Maximum features & $\sqrt{p}$\\
        & Splitting criterion & Friedman MSE\\
        \hline
        Support Vector Regressor & Kernel & Linear\\ 
        & Regularisation parameter & 2048\\
        & Gamma & 1e-7 \\
        & Epsilon & 0.5\\
        \hline
        Deep Neural Network & Number of layers & 8\\
        & Number of nodes for each layer & 64, 64, 32, 16, 8, 6, 4, 1 \\
        & Activation function for each layer & ReLU, except for output is linear \\
        & Loss function & MSE\\
        & Optimiser & Rectified Adam with LookAhead\\
        & Learning rate & 0.001\\
        & Kernel regulariser & L2 (0.001)\\
        & Dropout & 0.2\\
        \bottomrule
    \end{tabular}
    \caption{Hyperparameters value for Random Forest, Gradient Boosted Trees, Support Vector Regressor, and Deep Neural Network for all experiments.}
    \label{table:hyperparams}
\end{table}

The feature importance methods employed by each ML models are listed in Table~\ref{table:interpret-method}. For SHAP, we employ weighted k-means to summarise the data before estimating the values of SHAP. Each cluster is weighted by the number of points they represent. Using k-means to summarise the data has the advantage of lower computational cost but slightly decreasing the accuracy of SHAP values. However, we compare the SHAP values from data with and without k-means for several datasets and found the SHAP values to be almost identical.
 
 \begin{table}[htpb]
    \centering
    \begin{tabular}{ll}
        \toprule
        Models & Interpretability methods  \\
        \midrule
        Random Forest & Permutation Importance, SHAP \\
        \hline
        Gradient Boosted Trees & Permutation Importance, SHAP\\
        \hline
        Support Vector Regressor & Permutation Importance, SHAP\\
        \hline
        Deep Neural Network & Permutation Importance, SHAP, and Integrated Gradient\\
        \bottomrule
    \end{tabular}
    \caption{Interpretability methods employed by each ML model for feature importance fusion.}
    \label{table:interpret-method}
\end{table}

\subsection{Evaluation Metrics}
\label{sec:evaluation}
To evaluate the performance of the ensemble feature importance and also the different ensemble strategies we employ three different evaluation metrics namely, Mean Absolute Error (MAE), Root Mean Square Error (RMSE), and $R^{2}$.

\section{Results and Discussion}

\subsection{Single-Method Ensemble vs our Multi-Method Ensemble Framework}

\begin{figure}[htpb]
    \centering
    \includegraphics[width=1\textwidth]{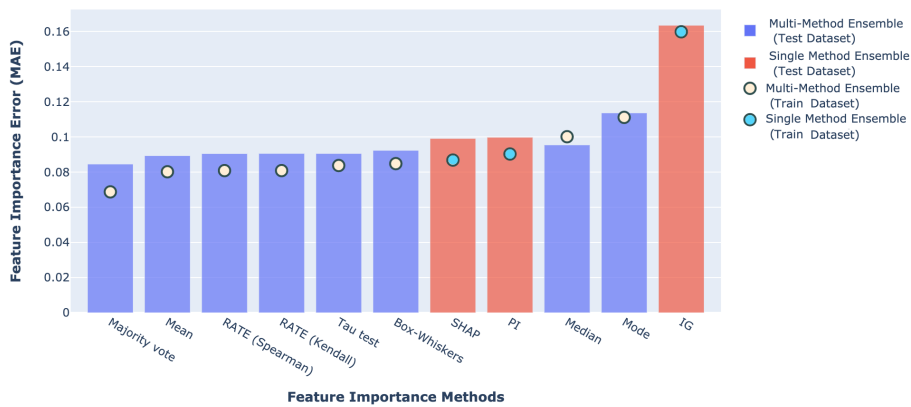}
    \caption{Average feature importance error between SME and our MME Framework with train and test dataset.}
    \label{fig:ens_indi}
\end{figure}

In order to compare results of SME and our MME Framework, Figure~\ref{fig:ens_indi} shows the average MAE of all SME and the multiple fusion implementations of our MME Framework across all datasets. Results using the RMSE and $R^{2}$ metrics are shown in the supplementary material. The ensemble method with the least feature importance error is our MME Framework using majority vote for fusion, followed by the MME Framework with mean and RATE. SME such as SHAP, PI and IG produce the worst results. The circles and bars in the figure represent the feature importance errors on the training and test datasets, respectively. The feature importance errors on the training dataset are slightly lower than errors on the test dataset. 

% \begin{figure}[H]
%     \centering
%     \includegraphics[width=1.0\textwidth]{Images/ensemble_percent.pdf}
%     \caption{Average feature importance error between the RATE, box-whiskers, Modified Thompson Tau test, median, mode, and majority vote ensemble methods against the mean ensemble strategy.}
%     \label{fig:ens_percent}
% \end{figure}

%%%%%%%%%%%%%%%%%%%%%%%%%%%%%%%%%%%%%%%%%%%%%%%%%%%%%%%%%%%%%%%%%%%%%%%%%%%%%%%%%%%%
%%%%%%%%%%%%%%%%%%%%%%%%%%%%%%%%%%%%%%%%%%%%%%%%%%%%%%%%%%%%%%%%%%%%%%%%%%%%%%%%%%%%

\subsection{Effect of Noise Level, Informative Level, and Number of Features on All Feature Importance}

\begin{figure}[htpb]
    \centering
    \includegraphics[width=1\textwidth]{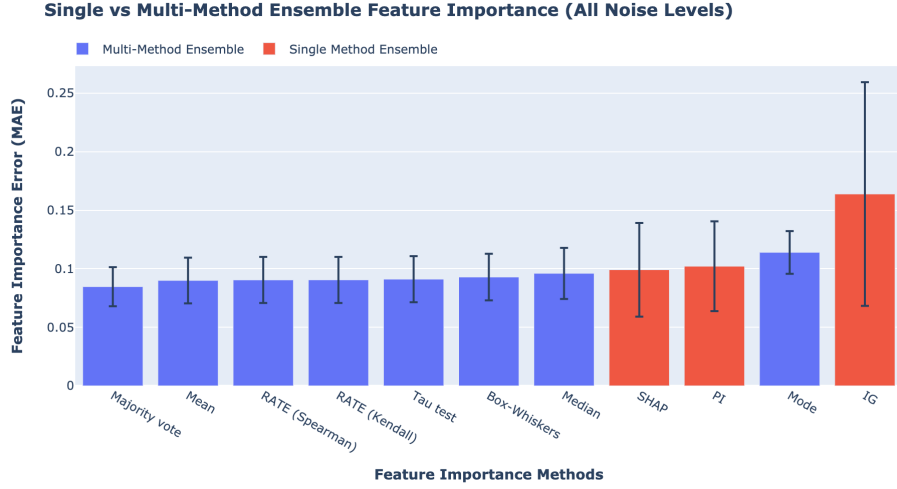}
    \caption{Effect of all noise level on all feature importance methods}
    \label{fig:all_noise}
\end{figure}

\begin{figure}[H]
    \centering
    \includegraphics[width=1.1\textwidth]{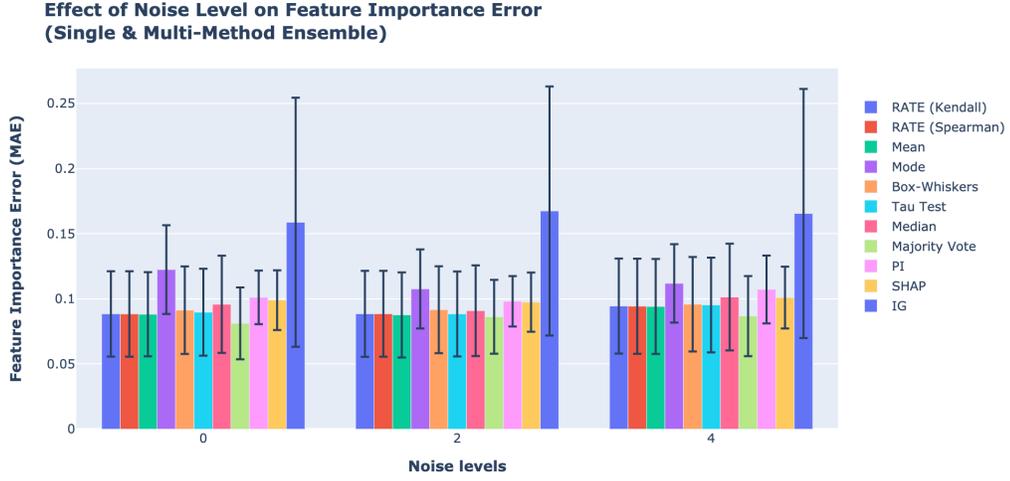}
    \caption{Effect of noise levels on ensemble feature importance.}
    \label{fig:noise_ensemble}
\end{figure}

Figure~\ref{fig:all_noise} shows the feature importance results of SME and the MME Framework with different fusion strategies averaged across three different noise levels in the data. RMSE and $R^{2}$ results are found in the supplementary material. The best performing ensemble method averaged across all noise levels is MME Framework using majority vote. MME Framework that uses majority vote outperforms the best SME method, SHAP, by 14.2\%. In addition, Table~\ref{table:summary-noise-level} and Figure~\ref{fig:noise_ensemble} show how the feature importance errors change for all SME and MME Framework methods as the noise level increases. In Table~\ref{table:summary-info-level} we observe that the MAE decreases marginally, from a noise level of 0 standard deviation to 2 standard deviations, and then it increases again 4 times the standard deviation. The addition of 2 standard deviation noise to the dataset improves the generalisation performance of ML models leading to lower errors~\cite{Bishop1995noise}. However, the feature importance errors increase when the noise in the dataset reaches 4 times the standard deviation, indicating that the noise level has negatively impacted ML models performance. Overall, however, the noise levels have little effect on the feature importance errors. Results also reveal that MME Framework with majority vote achieves the best feature importance estimates for our data.

\begin{table}[H]
    \centering
    \begin{tabular}{clrrr}
        \toprule
         && \multicolumn{3}{c}{Noise level (Standard deviation)}  \\
         \cline{3-5}
         &Models & 0 ($10^{-2}$) & 2 ($10^{-2}$) & 4 ($10^{-2}$) \\
        \midrule
        \multirow{3}{*}{\rotatebox[origin=c]{90}{SME}}&PI & 10.1$\pm$2.0 & 9.8$\pm$1.9 & 10.7$\pm$2.6  \\
        &SHAP & 9.8$\pm$2.2 & 9.7$\pm$2.2 & 10.0$\pm$2.3  \\
        &IG & 15.8$\pm$9.5 & 16.7$\pm$9.5 & 16.5$\pm$9.5 \\
        \midrule
        \multirow{8}{*}{\rotatebox[origin=c]{90}{MME}}&RATE (Kendall) & 8.8$\pm$3.2 & 8.8$\pm$3.2 & 9.4$\pm$3.6  \\
        &RATE (Spearman) & 8.8$\pm$3.2 & 8.8$\pm$3.2 & 9.4$\pm$3.6  \\
        &Median & 9.5$\pm$3.7 & 9.0$\pm$3.4 & 10.1$\pm$4.0  \\
        &Mean & 8.8$\pm$3.2 & 8.7$\pm$3.2 & 9.4$\pm$3.6  \\
        &Mode & 12.2$\pm$3.4 & 10.7$\pm$3.0 & 11.1$\pm$3.0  \\
        &Box-whiskers & 9.1$\pm$3.3 & 9.1$\pm$3.3 & 9.5$\pm$3.6  \\
        &Tau test & 8.9$\pm$3.3 & 8.8$\pm$3.2 & 9.5$\pm$3.6  \\
        &Majority vote & \textbf{8.1$\pm$2.7} & \textbf{8.6$\pm$2.8} & \textbf{8.6$\pm$3.0}  \\
        \bottomrule
    \end{tabular}
    \caption{Summary of feature importance MAE between different methods using SME and MME Framework for different noise level.}
    \label{table:summary-noise-level}
\end{table}

% \begin{figure}[H]
%     \centering
%     \includegraphics[width=1.2\textwidth]{Images/noise_pi.pdf}
%     \caption{Effect of noise levels on Permutation Importance.}
%     \label{fig:noise_pi}
% \end{figure}

% \begin{figure}[H]
%     \centering
%     \includegraphics[width=1.0\textwidth]{Images/noise_shap.pdf}
%     \caption{Effect of noise levels on SHAP.}
%     \label{fig:noise_shap}
% \end{figure}

% \begin{figure}[H]
%     \centering
%     \includegraphics[width=1.0\textwidth]{Images/noise_ig.pdf}
%     \caption{Effect of noise levels on Integrated Gradient.}
%     \label{fig:noise_ig}
% \end{figure}

%%%%%%%%%%%%%%%%%%%%%%%%%%%%%%%%%%%%%%%%%%%%%%%%%%%%%%%%%%%%%%%%%%%%%%%%%%%%%%%%%%%%
%%%%%%%%%%%%%%%%%%%%%%%%%%%%%%%%%%%%%%%%%%%%%%%%%%%%%%%%%%%%%%%%%%%%%%%%%%%%%%%%%%%%

\begin{figure}[H]
    \centering
    \includegraphics[width=1.0\textwidth]{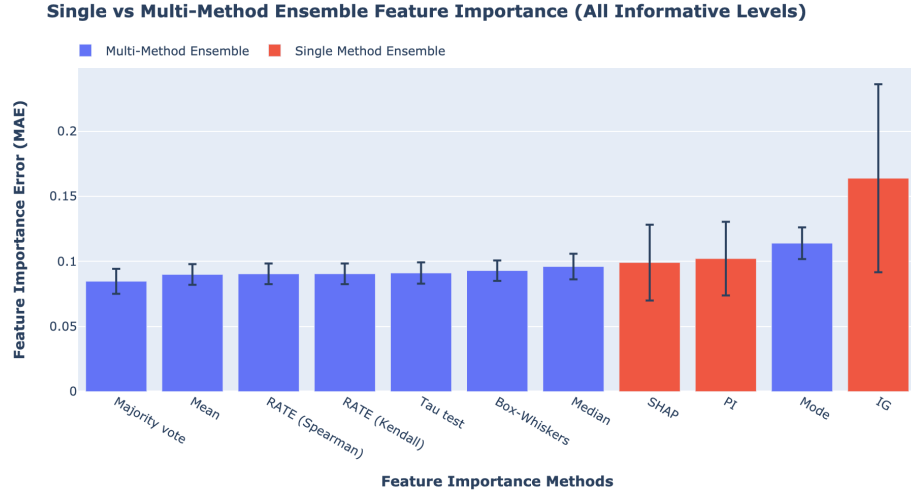}
    \caption{Effect of feature informative level on all ensemble feature importance methods.}
    \label{fig:all_info}
\end{figure}

Figure~\ref{fig:all_info} presents the feature importance MAE error for all SME and MME Framework methods. The best performing method is MME with majority vote followed by MME with mean and RATE. All MME Framework methods except for that with mode outperform SME by more accurately quantifying the feature importance. The error bar (standard deviation) for the effect of features informative levels in Figure~\ref{fig:all_info} has a smaller range than the effect of noise and the effect of number of features. The low standard deviation indicates that the feature informative levels explain most of the variances. From Table~\ref{table:summary-info-level} we observe that when 20\% of the features are informative, the best feature importance methods are MME Framework with Modified Thompson tau test and median, and they outperform the best SME method, SHAP, by 9.1\%. For 40\% features informative level, MME Framework with RATE (Kendall and Spearman) and mean have the lowest error, with 6.9\% improvement over SME with SHAP. For 60\% features informative level, MME Framework using RATE (Kendall and Spearman) have the best results, with a 4.1\% improvement over SME with SHAP. Furthermore, MME Framework with majority vote obtains the lowest error for both 80\% and 100\% features informative level. The best performing SME is using PI. MME with majority votes outperforms SME's best results by 25.4\% and 23.0\% for 80\% and 100\% features informative level, respectively.

\begin{table}[H]
    \centering
    \begin{tabular}{clrrrrr}
        \toprule
         && \multicolumn{5}{c}{Features informative level (\%)}  \\
         \cline{3-7}
         &Models &  20 ($10^{-2}$) & 40 ($10^{-2}$) & 60 ($10^{-2}$) & 80 ($10^{-2}$) & 100 ($10^{-2}$) \\
        \midrule
        \multirow{3}{*}{\rotatebox[origin=c]{90}{SME}}&PI & 2.7$\pm$0.4 & 6.7$\pm$0.8 & 11.2$\pm$2.0 & 13.8$\pm$1.0 & 16.5$\pm$1.3   \\
        &SHAP & 2.2$\pm$0.4 & 5.8$\pm$0.8 & 9.7$\pm$1.0 & 14.2$\pm$1.5 & 17.3$\pm$1.9   \\
        &IG & 6.3$\pm$7.2& 10.7$\pm$7.2 & 15.8$\pm$7.2 & 22.2$\pm$7.2 & 26.7$\pm$7.2   \\
        \midrule
        \multirow{8}{*}{\rotatebox[origin=c]{90}{MME}}&RATE (Kendall) &  2.1$\pm$0.4 & \textbf{5.4$\pm$1.1} & \textbf{9.3$\pm$1.5} & 12.3$\pm$1.6 & 15.9$\pm$3.0   \\
        &RATE (Spearman) &  2.1$\pm$0.5 & \textbf{5.4$\pm$1.1} & \textbf{9.3$\pm$1.5} & 12.3$\pm$1.6 & 15.9$\pm$3.0    \\
        &Median &  \textbf{2.0$\pm$0.6} & 5.9$\pm$1.4 & 10.2$\pm$1.8 & 13.0$\pm$2.4 & 16.7$\pm$3.5   \\
        &Mean &  2.1$\pm$0.5 & \textbf{5.4$\pm$1.1}  & 9.4$\pm$1.5 & 12.3$\pm$1.6 & 15.7$\pm$3.0  \\
        &Mode &  7.0$\pm$2.0 & 9.0$\pm$3.1  & 12.4$\pm$3.1 & 13.3$\pm$1.9 & 15.0$\pm$3.0  \\
        &Box-whiskers &  2.1$\pm$0.6 & 5.6$\pm$1.1 & 9.5$\pm$1.4 & 12.9$\pm$1.7 & 16.1$\pm$2.9   \\
        &Tau test & \textbf{2.0$\pm$0.6} & 5.6$\pm$1.2 & 9.5$\pm$1.5 & 12.4$\pm$1.7 & 15.7$\pm$3.0   \\
        &Majority vote &  3.1$\pm$0.9 & 6.5$\pm$1.5 & 9.4$\pm$1.6 & \textbf{10.3$\pm$2.1} & \textbf{12.7$\pm$3.4}   \\
        \bottomrule
    \end{tabular}
    \caption{Summary of feature importance MAE between different SME and our MME Framework for different percentage of informative level.}
    \label{table:summary-info-level}
\end{table}

\begin{figure}[H]
    \centering
    \includegraphics[width=1.2\textwidth]{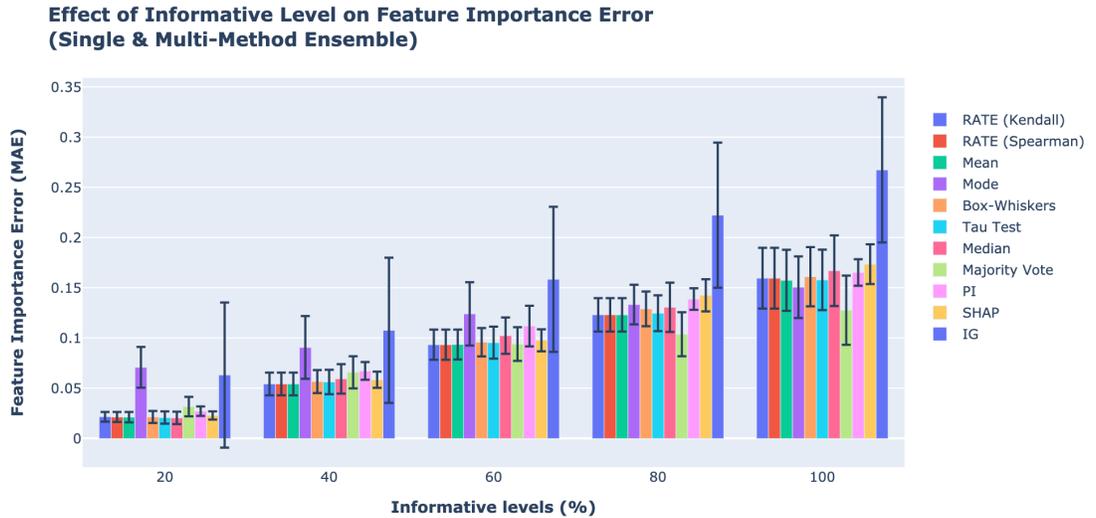}
    \caption{Effect of feature informative levels on ensemble feature importance methods.}
    \label{fig:info_ens}
\end{figure}

Figure~\ref{fig:info_ens} shows the relationship between the MAE of feature importance errors and the percentage of features informative level. As the percentage of feature informative level increases the feature importance errors increases. The higher number of contributing (non-zero) features to the output increases the difficulty of quantifying feature importance leading to higher errors.

%%%%%%%%%%%%%%%%%%%%%%%%%%%%%%%%%%%%%%%%%%%%%%%%%%%%%%%%%%%%%%%%%%%%%%%%%%%%%%%%%%%%
%%%%%%%%%%%%%%%%%%%%%%%%%%%%%%%%%%%%%%%%%%%%%%%%%%%%%%%%%%%%%%%%%%%%%%%%%%%%%%%%%%%%

\begin{figure}[H]
    \centering
    \includegraphics[width=1.0\textwidth]{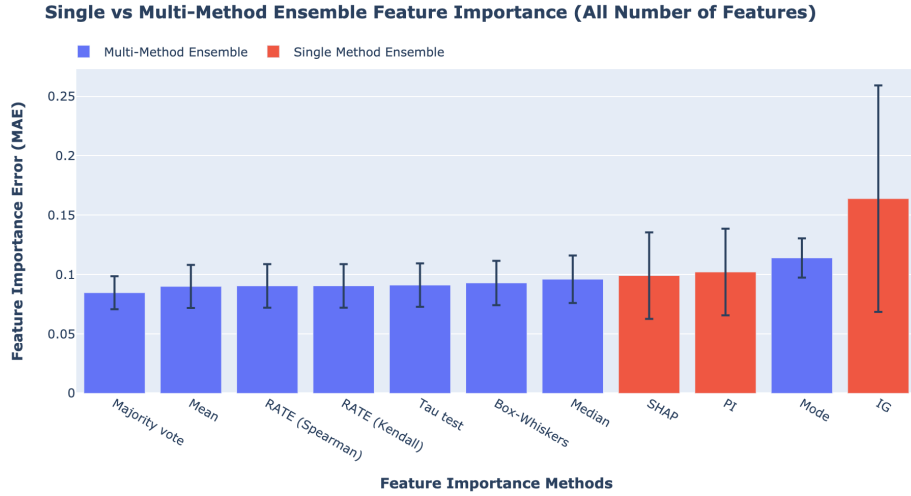}
    \caption{Effect of all number of features on all ensemble feature importance methods.}
    \label{fig:all_feat}
\end{figure}

 Figure~\ref{fig:all_feat} depicts the average of all SME and MME Framework with fusion methods across 20, 60, and 100 features. Similar to the effect of noise level and the percentage of informative features on feature importance errors, MME Framework with majority vote has the lowest error across all number of features.  The error bar for the effect of feature number in Figure~\ref{fig:all_feat} has a smaller range compared to the feature importance errors of the effect of noise but larger than the effect of number of features. From Table~\ref{table:summary-nfeat-level} we observe that when there are 20 and 40 features the method with the lowest feature importance errors is MME Framework with majority vote, and it outperforms the best SME method - SHAP by 8.2\% and 26.2\% respectively. For 100 features, MME Framework using mode is the most accurate method, and it outperforms the best SME method, PI by 20.5\%. For SME, PI has lower feature importance errors compared to SHAP as the number of informative features and features increases.

\begin{table}[H]
    \centering
    \begin{tabular}{clrrr}
        \toprule
         && \multicolumn{3}{c}{Number of features}  \\
         \cline{3-5}
         &Models & 20 ($10^{-2}$) & 60 ($10^{-2}$) & 100 ($10^{-2}$) \\
        \midrule
        \multirow{3}{*}{\rotatebox[origin=c]{90}{SME}}&PI & 7.5$\pm$1.8 &  10.8$\pm$2.0  &  12.2$\pm$2.4  \\
        &SHAP & 6.1$\pm$1.5 &  10.3$\pm$2.2  &  13.1$\pm$2.4  \\
        &IG & 15.8$\pm$9.5 &  16.1$\pm$9.5  &  17.1$\pm$9.5  \\
        \midrule
        \multirow{8}{*}{\rotatebox[origin=c]{90}{MME}}&RATE (Kendall) & 6.3$\pm$2.3 &  9.4$\pm$3.1  &  11.3$\pm$3.8  \\
        &RATE (Spearman) & 6.3$7\pm$2.3 &  9.4$\pm$3.1  &  11.3$\pm$3.8  \\
        &Median & 6.0$\pm$2.3 &  10.6$\pm$3.8  &  12.0$\pm$3.9  \\
        &Mean & 6.2$\pm$2.2 & 9.3$\pm$3.1  &  11.3$\pm$3.8  \\
        &Mode & 14.8$\pm$3.6 &  9.6$\pm$2.3  &  \textbf{9.7$\pm$2.4}  \\
        &Box-whiskers & 6.5$\pm$2.5 &  9.8$\pm$3.2  &  11.4$\pm$3.7  \\
        &Tau test & 6.1$\pm$2.4 &  9.7$\pm$3.2  &  11.4$\pm$3.7  \\
        &Majority vote & \textbf{5.6$\pm$1.9} &  \textbf{7.6$\pm$1.5}  &  12.1$\pm$3.3  \\
        \bottomrule
    \end{tabular}
    \caption{Summary of feature importance MAE between different SME and our MME Framework for different number of features.}
    \label{table:summary-nfeat-level}
\end{table}

\begin{figure}[H]
    \centering
    \includegraphics[width=1.1\textwidth]{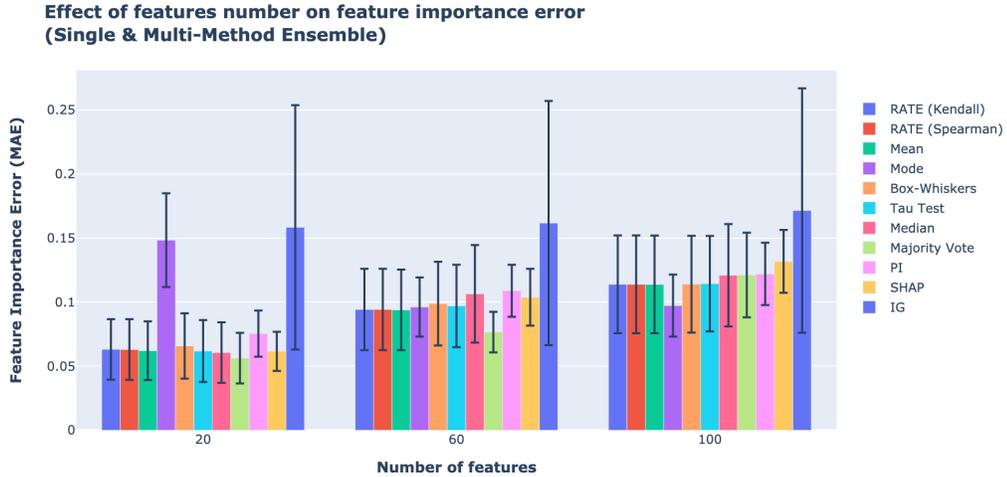}
    \caption{Effect of number of features on ensemble feature importance.}
    \label{fig:nfeat_ens}
\end{figure}

Figure~\ref{fig:nfeat_ens} shows the relationship between the MAE of feature importance and the number of features as they increase. Higher numbers of features increase the difficulty of quantifying feature importance accurately.

\label{sec:results}

\subsection{Discussion}
\label{sec:disscussion}

Overall, results show that our MME Framework outperforms SME for our case studies. In particular, the MME Framework with majority vote as an ensemble strategy for the three different combinations of factors: (i) noise level, (ii) feature informative level and (iii) number of features has achieved the best results. The robustness of our framework compared to SMEs becomes even more evident as the number of features and the number of informative features increases. Results also reveal that the noise in data does not affect the final feature importance estimates, as there are no significant changes as the noise increases. 

One advantage of using our MME Framework is that it avoids the worst-case scenario for feature importance estimates. This can also be a disadvantage as the best performing feature importance estimate is moderated by worse methods. However, for real-world, save critical systems where the ground truth is unknown, it is important not to rely solely on a single method and the potential bias it might produce. In scenarios where all feature importance determined by different methods disagree the best option for safety is to further investigate the reasons for disagreement.

For our experiments, we keep the hyperparameters of each ML models constant as their case-based optimisation is likely to affect feature importance estimates for different data characteristics. In the future, it is necessary to further investigate the interplay between parameter optimisation and feature importance. In addition, other factors such as covariate drift on the test dataset and imbalanced datasets should be included in the framework tests. 

\section{Conclusions and Future Work}
\label{sec:conclusions}

In this paper, we presented a novel framework to perform feature importance fusion using an ensemble of different ML models and different feature importance methods called MME. Overall, our framework performed better than traditional single model-agnostic approaches, where only one feature importance method with multiple ML models is employed. Additionally, we compared different ensemble strategies for our MME Framework such as measurements of central tendency such as median, mean, mode, and anomaly detection methods such as box-whiskers, Modified Thompson Tau test, and majority vote. In addition, we introduced a new ensemble strategy named RATE that combined rank correlation and majority vote. Furthermore, we studied the efficacy of MME Framework and SME on a combination of three different factors: (i) Noise level, (ii) percentage of informative features, and (iii) the number of features. We found that different noise level had minimal effect on feature importance, whereas the feature importance error increase with the percentage of informative features and number of features in data. For the case studies investigated, the MME Framework with majority vote is often the best performing method for all three data factors, followed by MME Framework with RATE and mean. When the number of features and the number of informative features are low, the performance of MME Framework is only marginally better than SME.

We showed that the MME Framework avoids the worst-case scenario in predicting feature importance. Despite its advantages, there are several shortcomings in the proposed method and our experimental approach that can be improved in future work. One improvement would be to include more feature importance methods and ML models into the MME Framework to increase the number of methods to investigate their impact on variance and in overall consensus of importance. Another possibility is to integrate the feature importance ensemble into a neural network to internally induce bias in the neural networks to focus on relevant features to decrease training time and improve accuracy. Finally, for the experimental aspect, several other factors that could potentially affect the feature importance estimates such as hyperparameter optimisations of ML models, imbalance of the datasets, the presence of anomalies in data, and the efficacy of the framework in real-world complex data should be investigated.

\section*{Acknowledgements}
This work is funded by the INNOVATIVE doctoral programme. The INNOVATIVE programme is partially funded by the Marie Curie Initial Training Networks (ITN) action (project number 665468) and partially by the Institute for Aerospace Technology (IAT) at the University of Nottingham.

\bibliographystyle{elsarticle-num}
\bibliography{ref}

\end{document}

\section{Supplementary results}
\label{sec:supplementary}
% \subsection{Multi-Method Ensemble in Contrast to Single-Method Ensemble (RMSE)}
% \begin{figure}[H]
%     \centering
%     \includegraphics[width=1.0\textwidth]{Images/rmse/rmse-fi.pdf}
%     \caption{Average feature importance error between SME and MME with train and test dataset (RMSE).}
%     \label{fig:ens_indi_rmse}
% \end{figure}

\subsection{Feature Importance Quantification on Train and Test Dataset (RMSE)}
\label{sec:rmse_train_test}
\begin{figure}[H]
    \centering
    \includegraphics[width=1.0\textwidth]{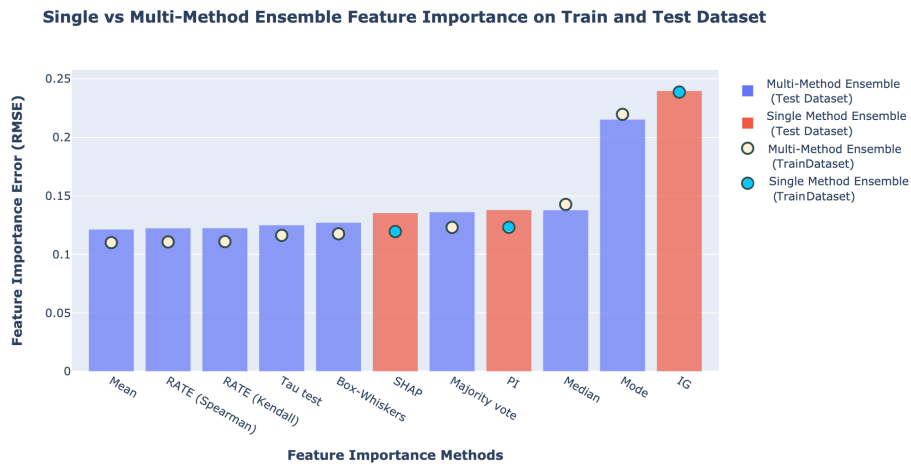}
    \caption{Average feature importance error between SME and MME with train and test dataset. (RMSE)}
    \label{fig:ens_indi_rmse}
\end{figure}

\subsection{Feature Importance Quantification on Train and Test Dataset ($R^{2}$)}
\label{sec:r2_train_test}
\begin{figure}[H]
    \centering
    \includegraphics[width=1.0\textwidth]{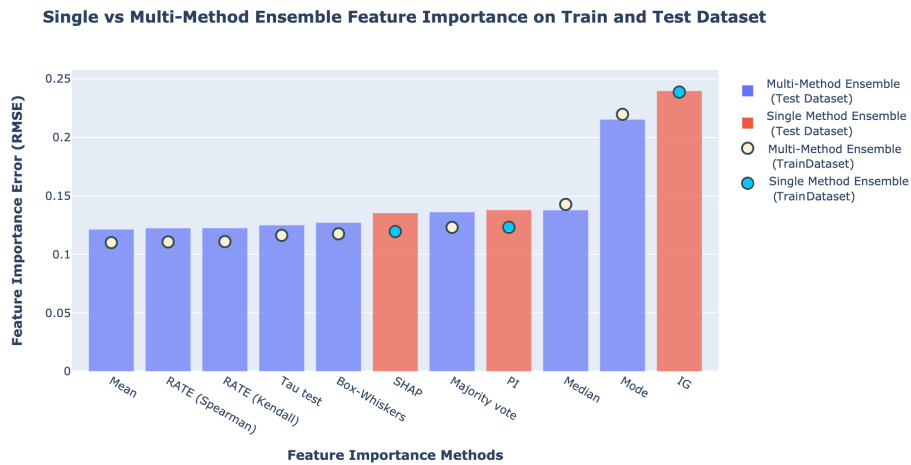}
    \caption{Average feature importance error between SME and MME with train and test dataset. (($R^{2}$))}
    \label{fig:ens_indi_r2}
\end{figure}

\subsection{Effect of Noise Level on All Feature Importance (RMSE)}

\begin{figure}[H]
    \centering
    \includegraphics[width=1.0\textwidth]{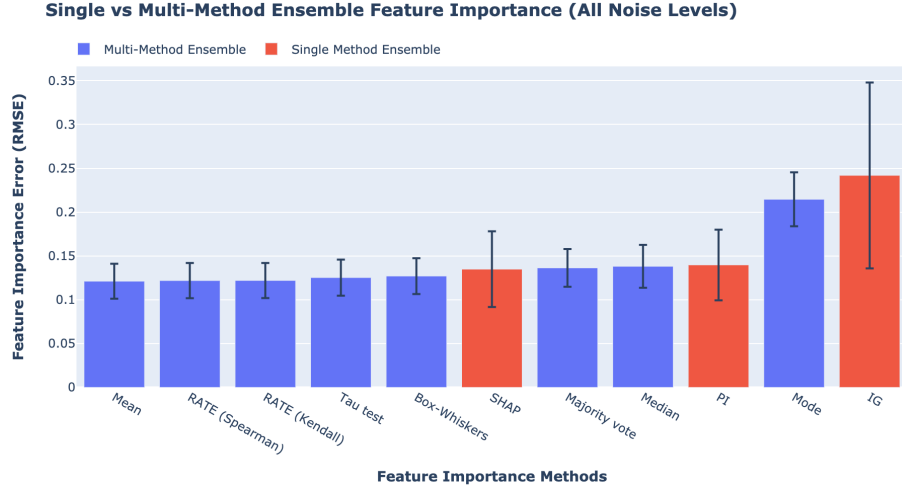}
    \caption{Effect of all noise level on all feature importance methods (RMSE)}
    \label{fig:all_noise_rmse}
\end{figure}

\begin{table}[H]
    \centering
    \begin{tabular}{clrrr}
        \toprule
         && \multicolumn{3}{c}{Noise level (Standard deviation)}  \\
         \cline{3-5}
         &Models & 0 ($10^{-2}$) & 2 ($10^{-2}$) & 4 ($10^{-2}$) \\
        \midrule
        \multirow{3}{*}{\rotatebox[origin=c]{90}{SME}}&PI & 13.8$\pm$2.1 & 13.4$\pm$2.0 & 14.5$\pm$2.7  \\
        &SHAP & 13.4$\pm$2.4 & 13.3$\pm$2.4 & 13.6$\pm$2.5  \\
        &IG & 23.1$\pm$10.6 & 25.0$\pm$10.5 & 24.3$\pm$10.6 \\
        \midrule
        \multirow{8}{*}{\rotatebox[origin=c]{90}{MME}}&RATE (Kendall) & 12.0$\pm$3.2 & 11.9$\pm$3.3 & 12.6$\pm$3.7  \\
        &RATE (Spearman) & 12.0$\pm$3.3 & 11.9$\pm$3.3 & 12.5$\pm$3.7  \\
        &Median & 13.8$\pm$4.1 & 13.8$\pm$3.2 & 14.3$\pm$4.6  \\
        &Mean & 12.0$\pm$3.2 & 11.8$\pm$3.3 & 12.5$\pm$3.7  \\
        &Mode & 22.8$\pm$5.4 & 19.9$\pm$5.3 & 21.5$\pm$5.1  \\
        &Box-whiskers & 12.5$\pm$3.4 & 12.6$\pm$3.4 & 12.9$\pm$3.7  \\
        &Tau test & 12.3$\pm$3.4 & 12.2$\pm$3.3 & 13.0$\pm$3.8  \\
        &Majority vote & 13.3$\pm$3.7 & 13.7$\pm$3.4 & 13.7$\pm$3.9  \\
        \bottomrule
    \end{tabular}
    \caption{Summary of feature importance RMSE between different SME and MME for different noise level.}
    \label{table:summary-noise-level-rmse}
\end{table}

\begin{figure}[H]
    \centering
    \includegraphics[width=1.0\textwidth]{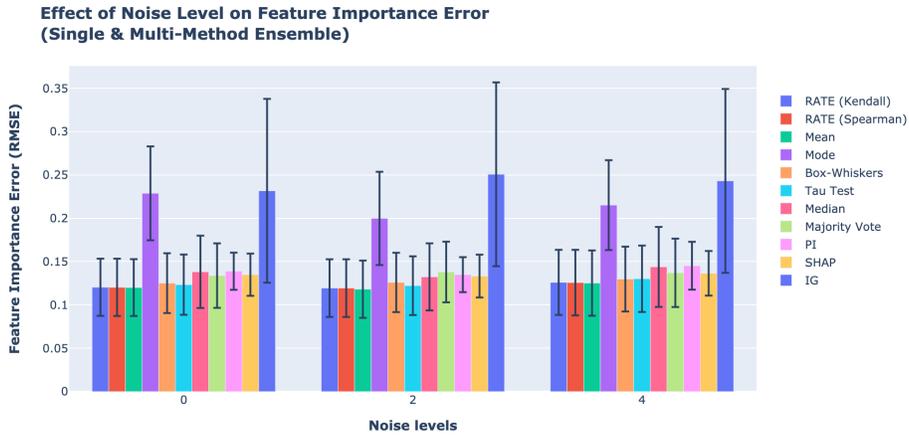}
    \caption{Effect of noise levels on ensemble feature importance. (RMSE)}
    \label{fig:noise_ensemble_rmse}
\end{figure}

\subsection{Effect of Noise Level on All Feature Importance ($R^{2}$)}

\begin{figure}[H]
    \centering
    \includegraphics[width=1.0\textwidth]{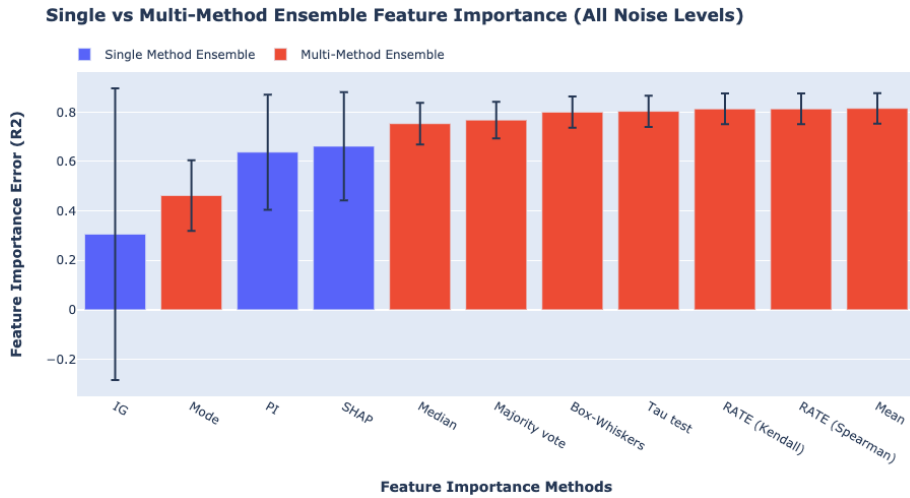}
    \caption{Effect of all noise level on all feature importance methods ($R^{2}$)}
    \label{fig:all_noise_r2}
\end{figure}

\begin{table}[H]
    \centering
    \begin{tabular}{clrrr}
        \toprule
         && \multicolumn{3}{c}{Noise level (Standard deviation)}  \\
         \cline{3-5}
         &Models & 0  & 2  & 4  \\
        \midrule0.
        \multirow{3}{*}{\rotatebox[origin=c]{90}{SME}}&PI & 0.63$\pm$0.11 & 0.67$\pm$0.10 & 0.60$\pm$0.17  \\
        &SHAP & 0.65$\pm$0.12 & 0.67$\pm$0.11 & 0.65$\pm$0.13  \\
        &IG & 0.29$\pm$0.59 & 0.28$\pm$0.59 & 0.34$\pm$0.59 \\
        \midrule
        \multirow{8}{*}{\rotatebox[origin=c]{90}{MME}}&RATE (Kendall) & 0.81$\pm$0.09 & 0.82$\pm$0.10 & 0.79$\pm$0.11  \\
        &RATE (Spearman) & 0.81$\pm$0.09 & 0.82$\pm$0.10 & 0.79$\pm$0.11  \\
        &Median & 0.75$\pm$0.15 & 0.78$\pm$0.12 & 0.72$\pm$0.15 \\
        &Mean & 0.81$\pm$0.09 & 0.82$\pm$0.10 & 0.79$\pm$0.11 \\
        &Mode & 0.39$\pm$0.26 & 0.52$\pm$0.25 & 0.46$\pm$0.22  \\
        &Box-whiskers & 0.80$\pm$0.10 & 0.80$\pm$0.10 & 0.79$\pm$0.11  \\
        &Tau test & 0.80$\pm$0.11 & 0.81$\pm$0.10 & 0.78$\pm$0.11  \\
        &Majority vote & 0.77$\pm$0.12 & 0.76$\pm$0.12 & 0.76$\pm$0.13  \\
        \bottomrule
    \end{tabular}
    \caption{Summary of feature importance $R^{2}$ between different SME and MME for different noise level.}
    \label{table:summary-noise-level-r2}
\end{table}

\begin{figure}[H]
    \centering
    \includegraphics[width=1.0\textwidth]{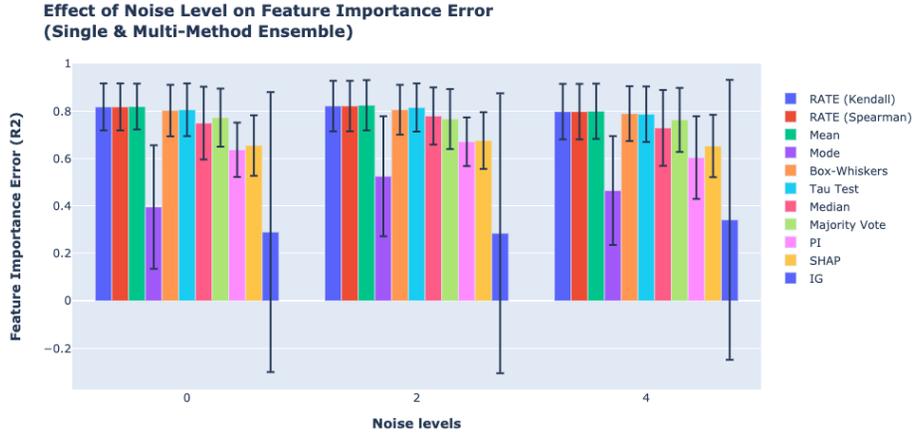}
    \caption{Effect of noise levels on ensemble feature importance. ($R^{2}$)}
    \label{fig:noise_ensemble_r2}
\end{figure}

\subsection{Effect Informative Level on All Feature Importance (RMSE)}

\begin{figure}[H]
    \centering
    \includegraphics[width=1.0\textwidth]{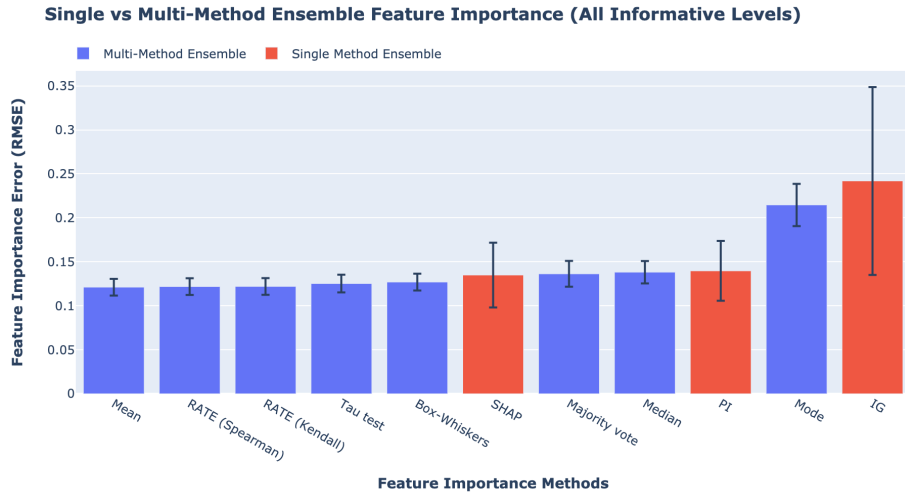}
    \caption{Effect of feature informative level on all ensemble feature importance methods. (RMSE)}
    \label{fig:all_info_rmse}
\end{figure}

\begin{table}[H]
    \centering
    \begin{tabular}{clrrrrr}
        \toprule
         && \multicolumn{5}{c}{Features informative level (\%)}  \\
         \cline{3-7}
         &Models &  20 ($10^{-2}$) & 40 ($10^{-2}$) & 60 ($10^{-2}$) & 80 ($10^{-2}$) & 100 ($10^{-2}$) \\
        \midrule
        \multirow{3}{*}{\rotatebox[origin=c]{90}{SME}}&PI & 6.0$\pm$0.9 & 11.1$\pm$1.2 & 15.6$\pm$2.3 & 17.4$\pm$1.2 & 19.5$\pm$1.4   \\
        &SHAP & 5.1$\pm$0.8 & 9.9$\pm$1.3 & 14.0$\pm$1.5 & 18.0$\pm$1.9 & 20.2$\pm$2.1   \\
        &IG & 14.9$\pm$10.6& 19.7$\pm$10.6 & 24.0$\pm$10.6 & 26.7$\pm$10.6 & 32.1$\pm$10.6   \\
        \midrule
        \multirow{8}{*}{\rotatebox[origin=c]{90}{MME}}&RATE (Kendall) &  5.0$\pm$1.1 & 9.2$\pm$1.7 & 13.2$\pm$2.1 & 15.3$\pm$1.9 & 18.1$\pm$3.1   \\
        &RATE (Spearman) &  5.0$\pm$1.1 & 9.2$\pm$1.7 & 13.2$\pm$2.1 & 15.3$\pm$1.9 & 18.1$\pm$3.1    \\
        &Median &  5.1$\pm$1.5 & 10.8$\pm$2.5 & 15.3$\pm$2.7 & 17.5$\pm$3.1 & 20.1$\pm$3.7   \\
        &Mean &  4.9$\pm$1.4 & 9.2$\pm$1.7  & 13.1$\pm$2.0 & 15.2$\pm$1.9 & 17.9$\pm$3.1  \\
        &Mode &  20.2$\pm$4.3 & 18.8$\pm$6.0  & 22.9$\pm$6.1 & 22.6$\pm$4.8 & 22.5$\pm$5.3  \\
        &Box-whiskers &  5.1$\pm$1.3 & 9.8$\pm$1.8 & 13.7$\pm$2.0 & 16.2$\pm$2.0 & 18.5$\pm$3.1   \\
        &Tau test & 5.0$\pm$1.4 & 9.7$\pm$1.8 & 13.7$\pm$2.2 & 15.8$\pm$2.0 & 18.2$\pm$3.1   \\
        &Majority vote &  8.0$\pm$2.4 & 12.2$\pm$3.1 & 15.3$\pm$2.8 & 15.1$\pm$3.2 & 16.9$\pm$4.4   \\
        \bottomrule
    \end{tabular}
    \caption{Summary of feature importance RMSE between different SME and MME for different percentage of informative level.}
    \label{table:summary-info-level-rmse}
\end{table}

\begin{figure}[H]
    \centering
    \includegraphics[width=1.2\textwidth]{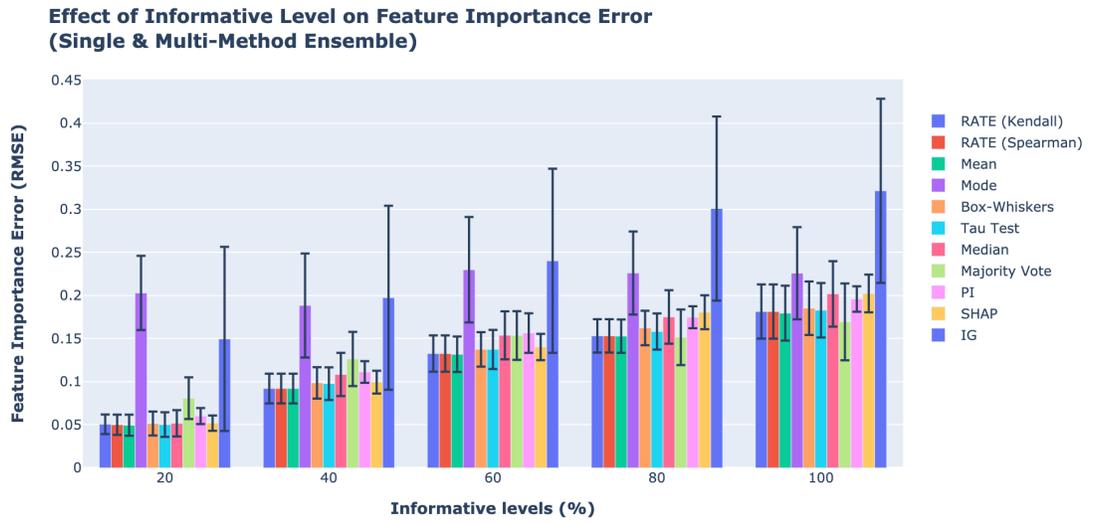}
    \caption{Effect of feature informative levels on ensemble feature importance methods. (RMSE)}
    \label{fig:info_ens_rmse}
\end{figure}

\subsection{Effect Informative Level on All Feature Importance ($R^{2}$)}

\begin{figure}[H]
    \centering
    \includegraphics[width=1.0\textwidth]{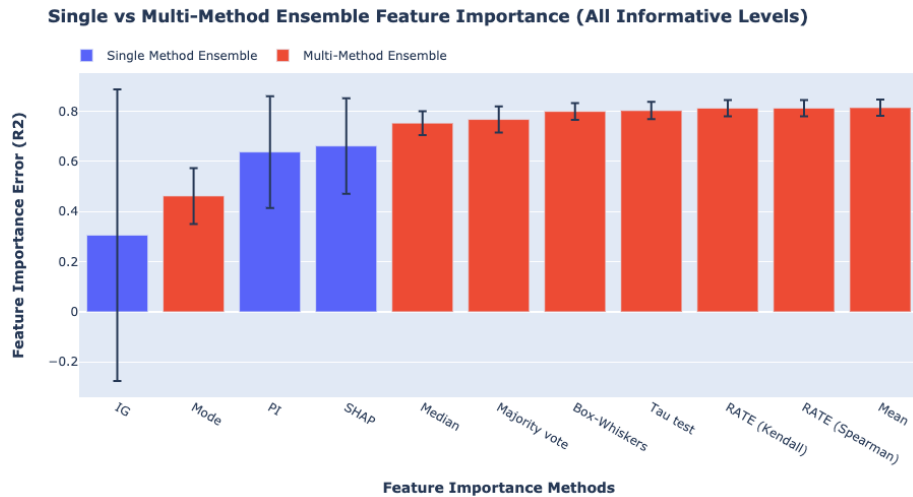}
    \caption{Effect of feature informative level on all ensemble feature importance methods. ($R^{2}$)}
    \label{fig:all_info_r2}
\end{figure}

\begin{table}[H]
    \centering
    \begin{tabular}{clrrrrr}
        \toprule
         && \multicolumn{5}{c}{Features informative level (\%)}  \\
         \cline{3-7}
         &Models &  20  & 40  & 60  & 80 & 100  \\
        \midrule
        \multirow{3}{*}{\rotatebox[origin=c]{90}{SME}}&PI & 0.90$\pm$0.02 & 0.79$\pm$0.04 & 0.63$\pm$0.16 & 0.57$\pm$0.06 & 0.27$\pm$0.12   \\
        &SHAP & 0.92$\pm$0.02 & 0.83$\pm$0.04 & 0.73$\pm$0.05 & 0.55$\pm$0.08 & 0.25$\pm$0.15   \\
        &IG & 0.65$\pm$0.58 & 0.58$\pm$0.58 & 0.44$\pm$0.58 & 0.17$\pm$0.58 & -0.32$\pm$0.58   \\
        \midrule
        \multirow{8}{*}{\rotatebox[origin=c]{90}{MME}}&RATE (Kendall) &  0.95$\pm$0.01 & 0.90$\pm$0.03 & 0.83$\pm$0.05 & 0.78$\pm$0.05 & 0.57$\pm$0.13   \\
        &RATE (Spearman) &  0.95$\pm$0.01 & 0.90$\pm$0.03 & 0.83$\pm$0.05 & 0.78$\pm$0.05 & 0.57$\pm$0.13    \\
        &Median &  0.95$\pm$0.02 & 0.86$\pm$0.05 & 0.77$\pm$0.07 & 0.70$\pm$0.09 & 0.46$\pm$0.19   \\
        &Mean &  0.95$\pm$0.01 & 0.90$\pm$0.03  & 0.83$\pm$0.05 & 0.78$\pm$0.05 & 0.58$\pm$0.13  \\
        &Mode &  0.37$\pm$0.22 & 0.57$\pm$0.28 & 0.49$\pm$0.24 & 0.52$\pm$0.19 & 0.34$\pm$0.29  \\
        &Box-whiskers &  0.95$\pm$0.02 & 0.89$\pm$0.03 & 0.82$\pm$0.05 & 0.75$\pm$0.06 & 0.56$\pm$0.13   \\
        &Tau test & 0.95$\pm$0.02 & 0.89$\pm$0.03 & 0.82$\pm$0.05 & 0.76$\pm$0.06 & 0.57$\pm$0.14   \\
        &Majority vote &  0.88$\pm$0.05 & 0.81$\pm$0.07 & 0.77$\pm$0.08 & 0.77$\pm$0.09 & 0.59$\pm$0.20   \\
        \bottomrule
    \end{tabular}
    \caption{Summary of feature importance $R^{2}$ between different SME and MME for different percentage of informative level.}
    \label{table:summary-info-level-r2}
\end{table}

\begin{figure}[H]
    \centering
    \includegraphics[width=1.2\textwidth]{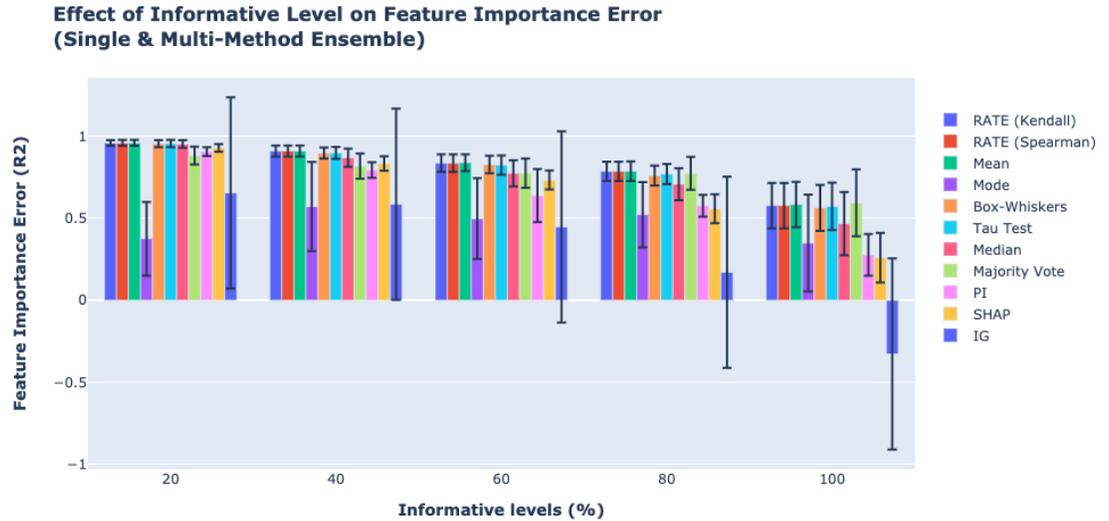}
    \caption{Effect of feature informative levels on ensemble feature importance methods. ($R^{2}$)}
    \label{fig:info_ens_r2}
\end{figure}

\subsection{Effect of Number of Features on All Feature Importance (RMSE)}

\begin{figure}[H]
    \centering
    \includegraphics[width=1.0\textwidth]{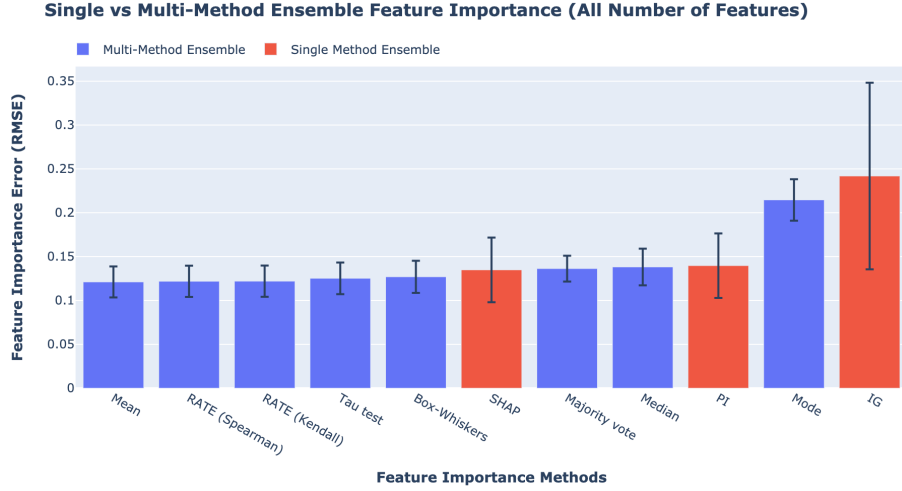}
    \caption{Effect of number of features on all ensemble feature importance methods. (RMSE)}
    \label{fig:all_nfeats_rmse}
\end{figure}

\begin{table}[H]
    \centering
    \begin{tabular}{clrrr}
        \toprule
         && \multicolumn{3}{c}{Number of features}  \\
         \cline{3-5}
         &Models & 20 ($10^{-2}$) & 60 ($10^{-2}$) & 100 ($10^{-2}$) \\
        \midrule
        \multirow{3}{*}{\rotatebox[origin=c]{90}{SME}}&PI & 10.3$\pm$1.9 &  14.9$\pm$1.9  &  16.5$\pm$2.4  \\
        &SHAP & 8.5$\pm$1.6 &  14.2$\pm$2.2  &  17.6$\pm$2.4  \\
        &IG & 23.6$\pm$10.6 &  24.0$\pm$10.6  &  24.7$\pm$10.6  \\
        \midrule
        \multirow{8}{*}{\rotatebox[origin=c]{90}{MME}}&RATE (Kendall) & 8.6$\pm$2.5 &  12.7$\pm$3.0  &  15.1$\pm$3.6  \\
        &RATE (Spearman) & 8.6$7\pm$2.5 &  12.7$\pm$3.0  &  15.1$\pm$3.6  \\
        &Median & 8.6$\pm$2.8 &  15.4$\pm$3.9  &  17.2$\pm$4.0  \\
        &Mean & 8.5$\pm$2.5 & 12.7$\pm$2.9  &  15.0$\pm$3.6  \\
        &Mode & 29.7$\pm$5.8 &  18.3$\pm$3.3  &  16.2$\pm$2.2  \\
        &Box-whiskers & 9.0$\pm$2.8 &  13.5$\pm$3.1  &  15.4$\pm$3.5  \\
        &Tau test & 8.5$\pm$2.7 &  13.4$\pm$3.0  &  15.5$\pm$3.5  \\
        &Majority vote & 8.3$\pm$2.3 &  13.0$\pm$3.1  &  19.4$\pm$3.3  \\
        \bottomrule
    \end{tabular}
    \caption{Summary of feature importance RMSE between different SME and MME for different number of features.}
    \label{table:summary-nfeat-level-rmse}
\end{table}

\begin{figure}[H]
    \centering
    \includegraphics[width=1.2\textwidth]{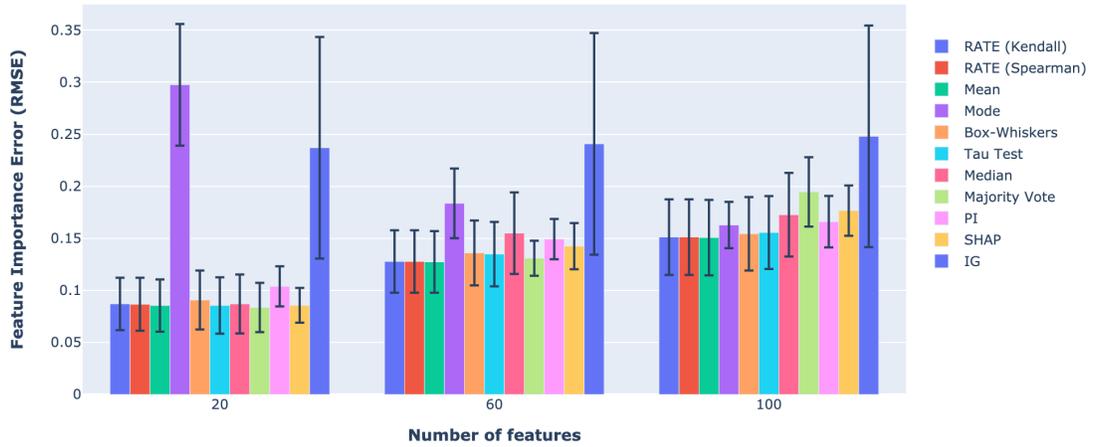}
    \caption{Effect of number of features on ensemble feature importance methods. (RMSE)}
    \label{fig:nfeats_ens_rmse}
\end{figure}

\subsection{Effect of Number of Features on All Feature Importance ($R^{2}$)}

\begin{figure}[H]
    \centering
    \includegraphics[width=1.0\textwidth]{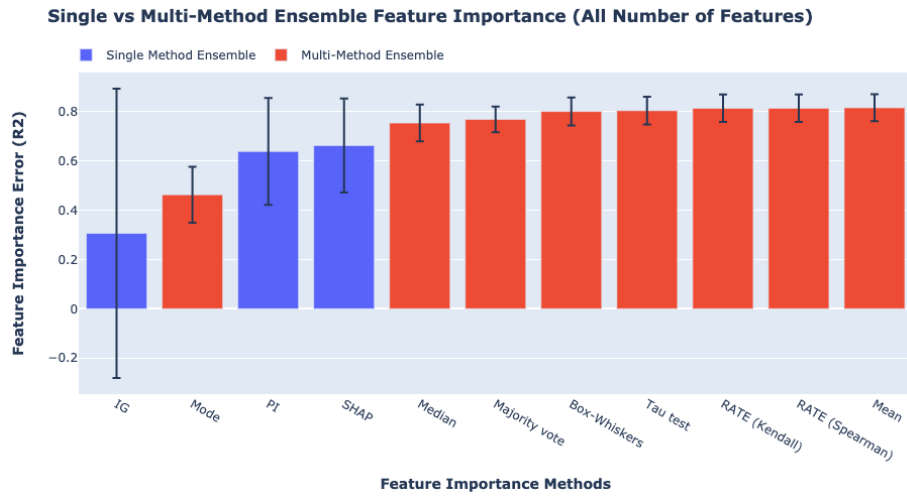}
    \caption{Effect of number of features on all ensemble feature importance methods. ($R^{2}$)}
    \label{fig:all_nfeats_r2}
\end{figure}

\begin{table}[H]
    \centering
    \begin{tabular}{clrrr}
        \toprule
         && \multicolumn{3}{c}{Number of features}  \\
         \cline{3-5}
         &Models & 20  & 60  & 100  \\
        \midrule
        \multirow{3}{*}{\rotatebox[origin=c]{90}{SME}}&PI & 0.81$\pm$0.07 &  0.60$\pm$0.11  &  0.50$\pm$0.16  \\
        &SHAP & 0.87$\pm$0.05 &  0.62$\pm$0.11  &  0.48$\pm$0.13  \\
        &IG & 0.34$\pm$0.58 &  0.32$\pm$0.58 &  0.24$\pm$0.58  \\
        \midrule
        \multirow{8}{*}{\rotatebox[origin=c]{90}{MME}}&RATE (Kendall) & 0.91$\pm$0.04 &  0.80$\pm$0.09  &  0.71$\pm$0.12  \\
        &RATE (Spearman) & 0.91$\pm$0.04 &  0.80$\pm$0.09  &  0.71$\pm$0.29  \\
        &Median & 0.91$\pm$0.04 &  0.71$\pm$0.14  &  0.63$\pm$0.16  \\
        &Mean & 0.91$\pm$0.04 & 0.80$\pm$0.09  &  0.71$\pm$0.12 \\
        &Mode & 0.09$\pm$0.29 &  0.60$\pm$0.14  &  0.68$\pm$0.08  \\
        &Box-whiskers & 0.90$\pm$0.04 &  0.78$\pm$0.09  &  0.70$\pm$0.13 \\
        &Tau test & 0.91$\pm$0.04 &  0.78$\pm$0.09  &  0.70$\pm$0.13  \\
        &Majority vote & 0.92$\pm$0.03 &  0.81$\pm$0.04  &  0.56$\pm$0.14  \\
        \bottomrule
    \end{tabular}
    \caption{Summary of feature importance $R^{2}$ between different SME and MME for different number of features.}
    \label{table:summary-nfeat-level-r2}
\end{table}

\begin{figure}[H]
    \centering
    \includegraphics[width=1.2\textwidth]{Images/r2/r2-nfeat-ens.pdf}
    \caption{Effect of number of features on ensemble feature importance methods. ($R^{2}$)}
    \label{fig:nfeats_ens_r2}
\end{figure}